\documentclass{article} 
\usepackage{iclr2026_conference,times}

\usepackage{amsmath,amsfonts,bm}

\def\eqref#1{equation~\ref{#1}}

\def\1{\bm{1}}

\def\vs{{\bm{s}}}

\DeclareMathAlphabet{\mathsfit}{\encodingdefault}{\sfdefault}{m}{sl}
\SetMathAlphabet{\mathsfit}{bold}{\encodingdefault}{\sfdefault}{bx}{n}

\usepackage[utf8]{inputenc} 
\usepackage[T1]{fontenc}    
\usepackage{hyperref}       
\usepackage{url}            
\usepackage{booktabs}       
\usepackage{amsfonts}       
\usepackage{nicefrac}       
\usepackage{microtype}      
\usepackage{multirow}
\usepackage{tikz}
\usepackage{xcolor}         
\usepackage{colortbl}
\usepackage[symbol]{footmisc} 
\usepackage{subcaption}
\usepackage{balance}
\usepackage{xspace}
\usepackage{amssymb}

\newcounter{iloop}
\usepackage{scalerel,stackengine,graphicx,xcolor}
\newcommand\openbigstar[1][0.7]{%
  \scalerel*{%
    \stackinset{c}{-.125pt}{c}{}{\scalebox{#1}{\color{white}{$\bigstar$}}}{%
      $\bigstar$}%
  }{\bigstar}%
}

\DeclareRobustCommand{\Stars}[1]{%
  \ensuremath{%
    \setcounter{iloop}{0}%
    \loop
      \ifnum\value{iloop}<\numexpr#1\relax
        \stepcounter{iloop}\bigstar
    \repeat
    \setcounter{iloop}{0}%
    \loop
      \ifnum\value{iloop}<\numexpr3-#1\relax
        \stepcounter{iloop}\openbigstar[.9]%
    \repeat
  }%
}

\makeatletter
\DeclareRobustCommand\onedot{\futurelet\@let@token\@onedot}
\def\@onedot{\ifx\@let@token.\else.\null\fi\xspace}

\newcommand{\app}{\raise.17ex\hbox{$\scriptstyle\sim$}}
 
\def\ie{\emph{i.e}\onedot} 
 
 \def\vs{\emph{vs}\onedot}

\makeatother

\definecolor{rowhl}{RGB}{229,239,249} 
\newlength\savewidth\newcommand\shline{\noalign{\global\savewidth\arrayrulewidth
  \global\arrayrulewidth 1pt}\hline\noalign{\global\arrayrulewidth\savewidth}}

\title{LightFusion: A Light-weighted, Double Fusion Framework for Unified Multimodal Understanding and Generation}

\author{
  \centerline{Zeyu Wang${^{1*}}$ \quad Zilong Chen${^{2,4*}}$ \quad Chenhui Gou${^{3,4*}}$ \quad \textbf{Feng Li}${^{4}}$ \quad \textbf{Chaorui Deng}${^{4}}$ } \vspace{.1em}\\
  \centerline{\textbf{Deyao Zhu}${^{4}}$ \quad \textbf{Kunchang Li}${^{4}}$ \quad \textbf{Weihao Yu}$^{4}$ \quad \textbf{Haoqin Tu}${^{1}}$ \quad \textbf{Haoqi Fan}$^{4}$ \quad \textbf{Cihang Xie}$^{1}$ \vspace{.3em}}\\
  \centerline{$^1$UC Santa Cruz \quad $^2$Tsinghua University \quad $^3$Monash University \quad $^4$ByteDance Seed} \vspace{.3em}
  \\
   \small \centerline{\textbf{Project Page}: \url{https://ucsc-vlaa.github.io/LightFusion/}}
  \centerline{\small $*$ Equal Contribution}
}

\iclrfinalcopy 
\begin{document}

\maketitle
\begingroup
\renewcommand\thefootnote{*}
\footnotetext{Equal contribution.}
\endgroup

\begin{abstract}
Unified multimodal models have recently shown remarkable gains in both capability and versatility, yet most leading systems are still trained from scratch and require substantial computational resources. In this paper, we show that competitive performance can be obtained far more efficiently by \textit{strategically} fusing publicly available models specialized for either generation or understanding.
Our key design is to retain the original blocks while additionally interleaving multimodal self-attention blocks throughout the networks. This \textit{double fusion mechanism} (1) effectively enables rich multi-modal fusion while largely preserving the original strengths of the base models, and (2) catalyzes synergistic fusion of high-level semantic representations from the understanding encoder with low-level spatial signals from the generation encoder. By training with only \app 35B tokens, this approach achieves strong results across multiple benchmarks: 0.91 on GenEval for compositional text-to-image generation, 82.16 on DPG-Bench for complex text-to-image generation, 6.06 on GEditBench, and 3.77 on ImgEdit-Bench for image editing. 
By fully releasing the entire suite of code, model weights, and datasets, we hope to support future research on unified multimodal modeling.
\end{abstract}

\section{Introduction}
In recent years, multimodal learning has witnessed a notable transition from specialized models to a new generation of unified multimodal models (UMMs). The central innovation of these models lies in their ability to natively couple language and vision, enabling them to understand both modalities and to generate text and images within a single, end-to-end modeling stack. Production-scale systems such as GPT-4o~\citep{hurst2024gpt} and Gemini 2.0 Flash~\citep{gemini20flash} have further demonstrated the promise of this paradigm, with impressive prompt adherence and dialogue-based image generation and manipulation capabilities, highlighting the potential of ``any-to-any'' paradigm.

Meanwhile, there is also a growing push inside the open-source research community to advance UMMs. The early works~\citep{team2024chameleon,wang2024emu3,zhou2024transfusion,wu2025janus,chen2025janus} adopt a single-stack transformer over mixed-modality sequences, but jointly optimizing autoregressive and diffusion objectives introduces fundamental training conflicts.
Subsequent works~\citep{deng2025emerging,wu2025omnigen2,wang2025skywork,wei2025skywork} decouples understanding and generation into separate pathways, easing optimization and improving performance, yet requiring large-scale training and substantial compute that limit accessibility.

Recent attempts to build UMMs efficiently---such as lightweight connectors that map final-layer representations from pretrained vision–language models (VLMs, \ie, used for understanding) to diffusion transformers (DiTs, \ie, used for generation) for conditional generation~\citep{pan2025transfer,chen2025blip3,lin2025uniworld}---are promising but remain incomplete, as performance and task generality are still limited.

\begin{figure*}
    \centering
    \includegraphics[width=\linewidth]{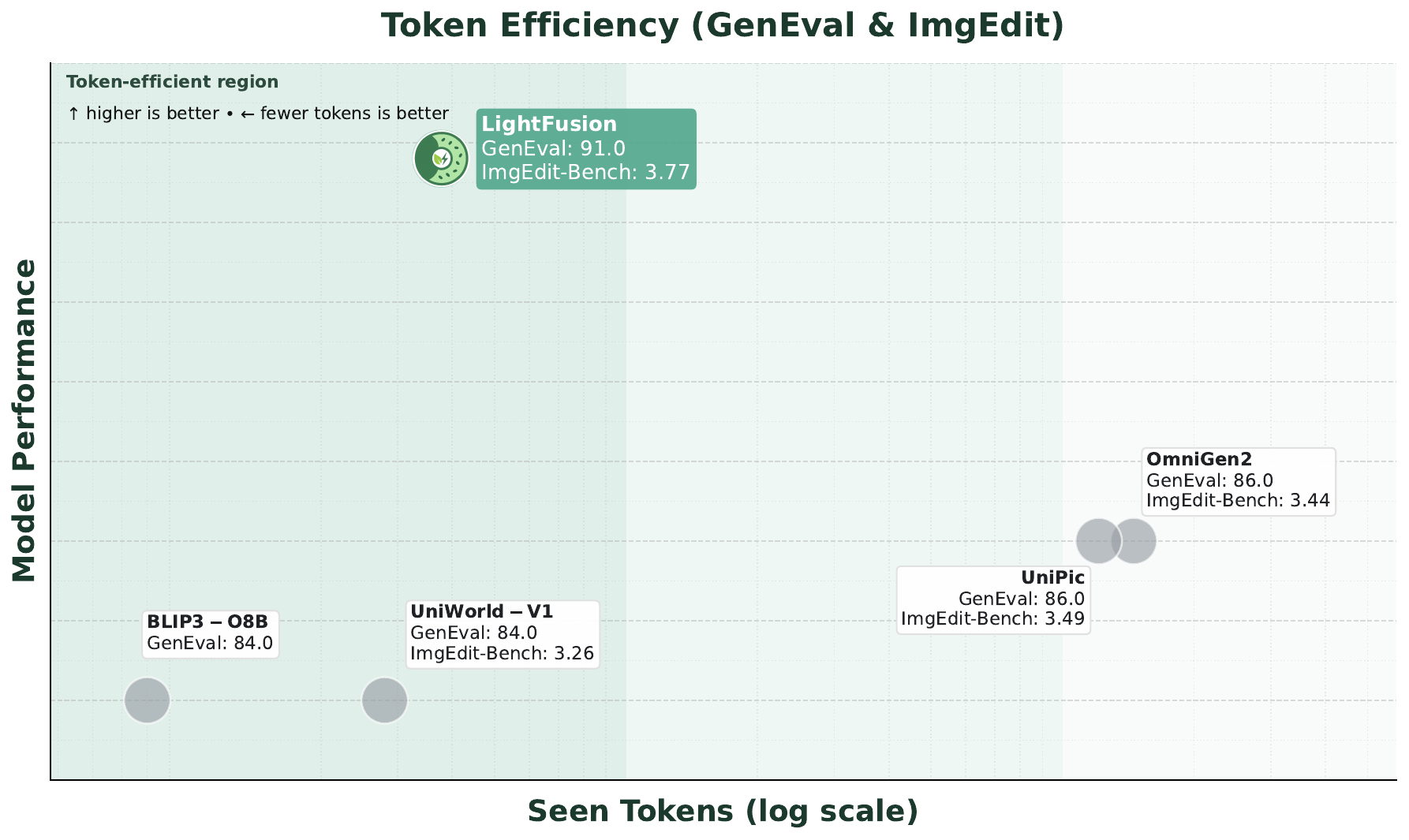}
    \caption{Token efficiency comparison on T2I and image editing benchmarks. Our LightFusion outperforms many leading unified models that uses significantly more tokens for training, showing great token efficiency. Note that we use our best estimate for the number of seen tokens of OmniGen2 and UniPic since their original training recipe is unclear to the public.}
    \label{fig:token_efficiency_merged}
    \vspace{-1em}
\end{figure*}

This work continues the pursuit of efficient UMM construction by introducing a novel and powerful fusion strategy.
At the core of our approach is a mechanism that tightly couples the VLM and DiT blocks of strong base models through zero-initialized multimodal self-attention layers. This design keeps the strong autoregressive and diffusion capabilities of the base models while enabling early, deep, and continuous cross‑modal interaction. Through controlled studies, we show that this strategy substantially outperforms the widely adopted ``Shallow Fusion'' approach~\citep{tang2025exploring}, which conditions the generator only on the final-layer output of the understanding branch~\citep{chen2025blip3,lin2025uniworld,wu2025omnigen2}. Furthermore, our design maintains two coordinated pathways that cleanly separate ViT tokens from VAE latents , allowing the model to naturally integrate high-level semantics with low-level visual signals. This synergy leads to more accurate, consistent, and faithful image editing behaviors.

We refer to this architecture as \textit{Double Fusion}, highlighting its dual role in simultaneously bridging (i) the understanding and generation branches and (ii) ViT-based and VAE-based visual representations. Together, these components enable efficient training and deliver strong performance while requiring substantially fewer tokens and compute than prior works, as illustrated in Figure ~\ref{fig:token_efficiency_merged}. The resulting model, which we call \textit{LightFusion}, is lightweight and modular, making it suitable for a wide range of applications and research settings.

LightFusion delivers state-of-the-art results across multiple benchmarks, including a GenEval score of 0.91 for compositional text-to-image generation, a DPG-Bench score of 82.16 for complex text-to-image generation, and scores of 6.06 on GEditBench and 3.77 on ImgEdit-Bench for image editing. Remarkably, trained on only \app 35B seen tokens, LightFusion achieves performance on par with, or even surpassing, leading models like UniPiC and OmniGen2, which were trained with orders of magnitude more tokens. These results highlight LightFusion’s efficiency and suggest new directions for the design of future UMM architectures. 

\section{Related Works}
\textbf{Text-to-Image Generation.} Diffusion models~\citep{ho2020denoising,song2020denoising} have emerged as the dominant paradigm for open-domain image synthesis, surpassing GANs~\citep{goodfellow2020generative} with improved stability and semantic fidelity. Seminal works, such as Stable Diffusion and its successors~\citep{rombach2022high,podell2023sdxl,esser2024scaling}, DALL·E~\citep{ramesh2022hierarchical}, and Imagen~\citep{imagen3}, demonstrated the power of large-scale pretraining and latent denoising for high-resolution, text-aligned generation. 
More recently, flow-matching models~\citep{lipman2022flow} have been proposed as a powerful alternative to diffusion, modeling vector fields between noise and data distributions, and have achieved strong results in large-scale systems~\citep{flux1dev,wan2025}.

\textbf{Image Editing.} 
Image editing has been widely studied as an extension of text-to-image generation. InstructPix2Pix~\citep{brooks2023instructpix2pix} pioneered the supervised paradigm based on \textit{\{instruction, source image, target image\}} triplets. Subsequent works improved this approach by diversifying training data and improving quality~\citep{zhang2023magicbrush,hui2025hqedit,wei2024omniedit,ye2025imgedit}, or by incorporating stronger control signals~\citep{tan2024ominicontrol,liu2025step1x,zhang2025context}. 
More recent frameworks such as OmniGen~\citep{xiao2025omnigen} and UniReal~\citep{chen2025unireal} unify common image generation tasks by processing various conditional inputs with modified attention mechanisms, demonstrating the potential of a single model for multiple generation tasks.

\textbf{Unified Multimodal Models.} UMMs have rapidly gained attention for their ability to support both understanding and generation, exhibiting strong generalization across visual understanding, generation, editing, and many other downstream tasks. Early UMM works~\citep{team2024chameleon,wang2024emu3,zhou2024transfusion,wu2025janus,chen2025janus,xie2024show} employed a single Transformer backbone to process interleaved image and text tokens. Subsequent work introduced separate understanding and generation branches, yielding improved performance while easing optimization towards two distinct tasks~\citep{shi2024lmfusion,deng2025emerging,wu2025omnigen2,wei2025skywork}. Nonetheless, these models generally depend on extensive large-scale pre-training over massive image–text corpora, which constrains their accessibility and limit the broader exploration of UMM researches.

Another line of works, such as BLIP3-o~\citep{chen2025blip3}, MetaQueries~\citep{pan2025transfer} and UniWorld~\citep{lin2025uniworld}, aims to build UMMs efficiently by coupling frozen multimodal LLMs with trainable diffusion decoders. Yet, existing schemes typically rely on shallow or late-stage fusion, restricting the expressiveness and depth of cross-modal alignment.

Our LightFusion model advances this direction through a double-fusion architecture. It tightly couples a powerful VLM with a DiT-based generator, enabling information to flow richly across modalities. This dual pathway design also naturally and effectively integrates high-level ViT features with low-level VAE latents, yielding a more coherent and semantically grounded generation pipeline. 

Notably, the double-fusion structure bears similarity to the modality-specialized mixture-of-experts designs in LMFusion~\citep{shi2024lmfusion} and BAGEL~\citep{deng2025emerging}, with drastically different underlying requirements. Those models initialize the generation branch with duplicated LLM weights, demanding orders of magnitude more seen tokens and extensive private training data. In contrast, LightFusion trains publicly available fused base models with only public data, achieving strong performance in an extremely lightweight and accessible training regime.

\section{Methods}
\subsection{Model Architecture}
\begin{figure*}
    \centering
    \includegraphics[width=0.9\linewidth]{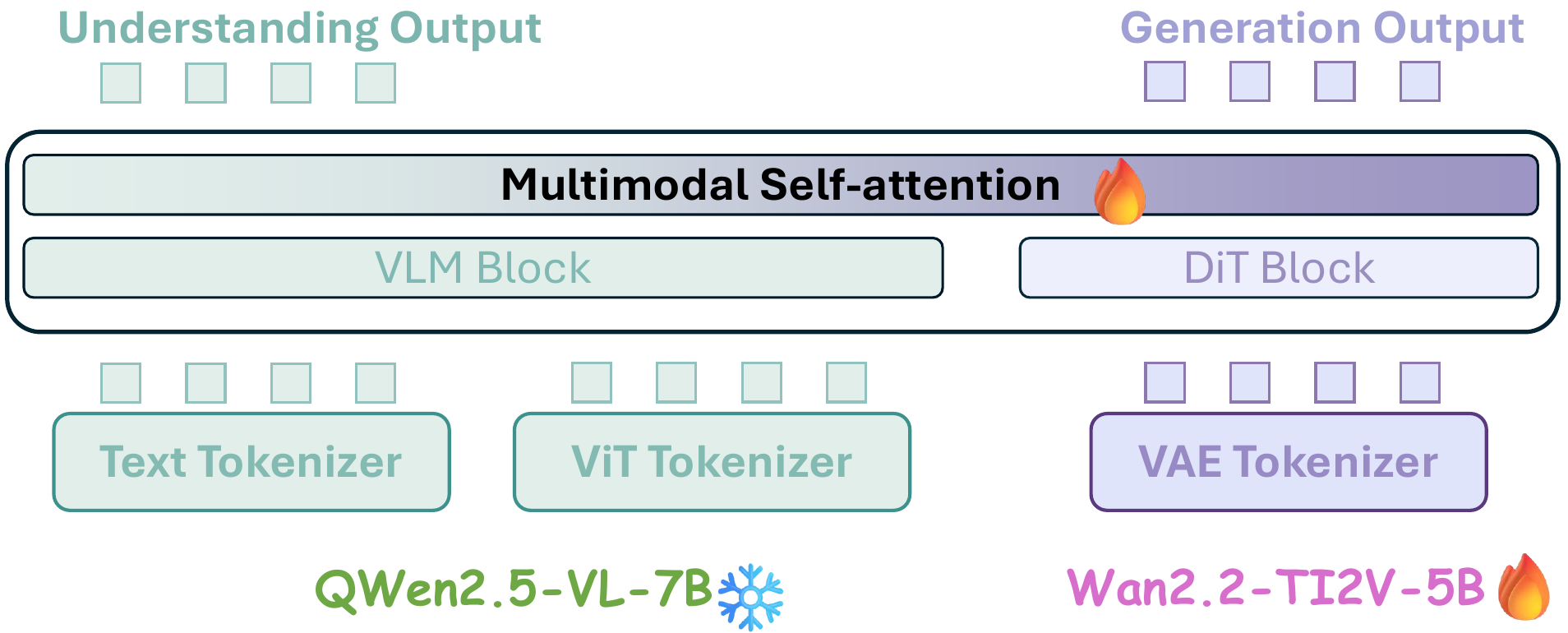}
    \caption{Overview of the LightFusion architecture. Text and ViT tokens (understanding pathway) and VAE tokens (generation pathway) are processed by pre-trained VLM and DiT blocks, respectively. At each layer, a zero-initialized multimodal self-attention module enables cross-modal interactions without altering the original model architectures.}
    \label{fig:arch}
    \vspace{-1em}
\end{figure*}

The core design of LightFusion is a Double Fusion mechanism. Specifically, as illustrated in Figure~\ref{fig:arch}, this design interleaves multimodal self-attention blocks across VLM (used for understanding) and DiT (used for generation) blocks. To fully leverage the strength of open-source models with extensive pre-training and post-training, we employ QWen2.5-VL-7B~\citep{bai2025qwen2} for the understanding pathway and Wan2.2-TI2V-5B~\citep{wan2025} for the generation pathway. Note that given QWen2.5-VL-7B has slightly fewer layers than Wan2.2-TI2V-5B (two layers fewer), its final-layer output is reused as input to the last two multimodal self-attention blocks.

In this design, the VLM blocks process understanding tokens (\ie, text and ViT tokens), while the DiT blocks operate on generation tokens (\ie, VAE tokens). The multimodal self-attention blocks span all token types, enabling rich cross-modal interactions. To facilitate this, we adopt the generalized causal attention mechanism from BAGEL~\citep{deng2025emerging}, allowing tokens from different modalities and tokenizers to attend to one another. Importantly, each multimodal self-attention block is zero-initialized, ensuring that the feature distributions of the VLM and DiT remain intact at the start of training, thereby preserving their strong autoregressive and denoising capabilities. For image editing tasks, both ViT and VAE tokens are extracted from the source image and provided as conditioning signals. The ViT tokens are extracted using the QWen2.5-VL vision encoders. The VAE tokens are extraced by the 3D causal VAE from Wan2.2-TI2V-5B~\citep{wan2025}, which provides 16$\times$ spatial compression and 4$\times$ temporal compression.

We expect this design offer several key advantages:
\begin{itemize}
\item \textbf{Seamless model integration}. The framework incorporates powerful pre-trained VLM and DiT models without altering their architectures, offering a straightforward and generalizable method for fusing publicly available models into a unified multimodal system. In line with prior findings~\citep{tang2025exploring}, we observe that this ``deep fusion'' strategy consistently outperforms ``shallow fusion'' approaches while delivering superior token efficiency.
\item \textbf{Dual-pathway visual representation}. By naturally integrating ViT tokens (high-level semantics) and VAE tokens (low-level signals), the architecture achieves precise and consistent image editing, effectively balancing global understanding with fine-grained detail.
\item \textbf{Information-preserving multimodal interaction}. Leveraging hidden states from all understanding and generation layers within the multimodal self-attention avoids compressing conditioning inputs into a fixed-length representation~\citep{pan2025transfer,chen2025blip3,wei2025skywork}, ensuring rich, loss-free cross-modal interactions.
\end{itemize}

\subsection{Dataset}
Our training corpus comprises  \app 45 million samples, encompassing both text-to-image generation and image-editing tasks. We source data from publicly available datasets, including BLIP-3o~\citep{chen2025blip3}, Civitai~\citep{civitai}, OmniGen~\citep{xiao2025omnigen}, OmniEdit~\citep{wei2024omniedit}, GPT-IMAGE-EDIT-1.5M~\citep{wang2025gpt}, and UniWorld-V1~\citep{lin2025uniworld}, supplemented by a synthetic self-curated dataset of \app 4.5 million samples. 
Notably, we apply a VLM to refine the editing instructions of public image editing data conditioned on source–target pairs, thereby enhancing instruction precision. 

\subsection{Training}
We adopt NaViT-style image processing to preserve native aspect ratios~\citep{dehghani2023patch}, constraining inputs to a minimum short side of 512 pixels and a maximum long side of 1024 pixels, thereby improving generation quality. ViT tokens are extracted from input sizes ranging from a minimum short side of 224 pixels to a maximum long side of 532 pixels. The LightFusion model is trained for 70K steps using the AdamW optimizer, with 2K warmup steps and a fixed learning rate of 0.00003. The sequence length is configured between 16,384 (minimum) and 20,480 (maximum) tokens. To enable classifier-free guidance, text tokens, VAE tokens, and ViT tokens are randomly dropped with probabilities of 0.1, 0.1, and 0.5, respectively. The understanding branch is frozen during the entire training time to preserve the strong understanding ability of QWen2.5-VL-7B.

We divide the full training process into three stages. In the first stage, a large proportion of common T2I data and a small proportion of high-quality T2I and editing data are used. In the second and third stages, we progressively increase the ratios of high-quality T2I and editing data, respectively. In practice, this staged setup proves beneficial for improving both text-to-image generation and image-editing performance.

\section{Experiments}
In this section, we evaluate the performance of LightFusion across a diverse set of image understanding, generation, and editing tasks.  
Sec.~\ref{sec:visual_und} presents the visual understanding results. Text-to-image generation results are reported in Sec.~\ref{sec:t2i_gen}, followed by a focus on image editing in Sec.~\ref{sec:image_editing}. Sec.~\ref{sec:ablation_study} presents our ablation study, analyzing key design choices of the model. Our experiment results show that LightFusion efficiently achieves strong performances over a wide spectrum of tasks and benchmarks, demonstrating the efficacy of our double fusion approach.

\begin{table*}[t]
\centering
\caption{Comparison of different models across understanding, generation, editing, and in-context Generation tasks. $\dagger$ refers to the methods using LLM rewriter. For UMMs, $\Stars{1}$ refers to only model checkpoints and evaluation code being released; $\Stars{2}$ refers to only model checkpoints and training/evaluation code being released; $\Stars{3}$ refers to the full suite of all the \{model, data, code\} being released.}
\vspace{-.5em}
\resizebox{\linewidth}{!}{
\begin{tabular}{l c c ccc cc cc}
\toprule
\multirow{2}{*}{Model} & \multirow{2}{*}{\# Params} & \multirow{2}{*}{Openness} &
\multicolumn{3}{c}{Understanding} & \multicolumn{2}{c}{Image Generation} & \multicolumn{2}{c}{Image Editing} \\
\cmidrule(lr){4-6}\cmidrule(lr){7-8}\cmidrule(lr){9-10}
 &  &  & MMB & MMMU & MM-Vet & GenEval & DPG & ImgEdit & GEdit-EN \\
\midrule
LLaVA-1.5        & --  & --           & 36.4 & 67.8 & 36.3 & --   & --    & --   & --   \\
LLaVA-NeXT       & --  & --             & 79.3 & 51.1 & 57.4 & --   & --    & --   & --   \\
\midrule
SDXL             & --  & --             & --   & --   & --   & 0.55 & 74.7  & --   & --   \\
SD3-medium       & -- & --              & --   & --   & --   & 0.62 & 84.1  & --   & --   \\
FLUX.1-dev       & -- & --              & --   & --   & --   & 0.66 & 84.0  & --   & --   \\
\midrule
Instruct-P2P     & -- & --             & --   & --   & --   & --   & --    & 1.88 & 3.68 \\
MagicBrush       & --  & --             & --   & --   & --   & --   & --    & 1.90 & 1.86 \\
AnyEdit          & --  & --             & --   & --   & --   & --   & --    & 2.45 & 3.21 \\
Step1X-Edit      & --  & --             & --   & --   & --   & --   & --    & 3.06 & 6.70 \\
IC-Edit          & -- & --             & --   & --   & --   & --   & --    & 3.05 & 4.84 \\
\midrule
\multicolumn{10}{l}{\bf \large \textit{\textcolor{orange}{Unified models}}}   \\
Janus-Pro        & --  & \Stars{1}             & 75.5 & 36.3 & 39.8 & 0.80 & 84.19 & --   & --   \\
Emu3             & --  & \Stars{1}             & 58.5 & 31.6 & 37.2 &0.66$^\dagger$ & 80.60 & -- & -- \\
UniPic           & 1.5B         & \Stars{1}    & -- & --  & --  & 0.86 & 85.50 & 3.49 & 5.83  \\
UniPic 2.0       & 7B + 2B      & \Stars{1}    & 83.5  & 58.6  & 67.1 & 0.90$^\dagger$ & 83.79 & 4.06 & 7.10  \\
Ovis-U1          & 2.4B + 1.2B  & \Stars{1} & 77.8 & 51.1 & 66.7 & 0.89 & 83.72 & 4.00 & 6.42  \\
MetaQuery-XL     & 7B + 1.6B    & \Stars{2} & 83.5 & 58.6 & 66.6 & 0.80$^\dagger$ & 82.05 & -- & --  \\
Show-o2          & 7B  & \Stars{2}             & 79.3 & 48.9 & --   & 0.76 & 86.14 & --   & --   \\
OmniGen          & 3.8B     & \Stars{2} & --   & --   & --   & 0.68 & 81.16 & 2.96 & 5.06 \\
OmniGen2         & 3B + 4B  & \Stars{2} & 79.1 & 53.1 & 61.8 &0.86$^\dagger$ & 83.57 & 3.44 & 6.42 \\
BAGEL            & 7B + 7B  & \Stars{2} & 85.0 & 55.3 & 67.2 &0.88$^\dagger$ & 85.07 &3.20 & 6.52 \\
\midrule
BLIP3-o 4B       & 3B + 1.4B  & \Stars{3} & 78.6 & 46.6 & 60.1 & 0.81$^\dagger$ & 79.36 & -- & -- \\
BLIP3-o 8B       & 7B + 1.4B  & \Stars{3} & 83.5 & 58.6 & 66.6 & 0.84$^\dagger$ & 81.60 & -- & -- \\
UniWorld-V1      & 7B + 12B   & \Stars{3} & 83.5 & 58.6 & 67.1 & 0.84$^\dagger$ & 81.38 & 3.26 & 4.85 \\
\rowcolor{blue!10}
LightFusion & 7B + 5B + 3B  & \Stars{3} & 83.5 & 58.6 &67.1 & 0.91$^\dagger$ &82.16 &3.77 &6.06 \\
\bottomrule
\end{tabular}
}
\label{tab:overall_results}
\end{table*}

\subsection{Visual Understanding}
\label{sec:visual_und}
By freezing the understanding branch, our model preserves the strong multimodal reasoning capabilities of the pre-trained QWen2.5-VL-7B. As reported in Table \ref{tab:overall_results}, LightFusion attains competitive performances of 83.5 on MMBench~\citep{liu2024mmbench}, 58.6 on MMMU~\citep{yue2024mmmu}, and 67.1 on MM-Vet~\citep{yu2023mm}. This architectural choice aligns with recent state-of-the-art designs, such as UniWorld-V1 \citep{lin2025uniworld}, OmniGen2 \citep{wu2025omnigen2}, and UniPiC 2.0 \citep{wang2025skywork}, which similarly prioritize maintaining robust understanding performance. As a result, LightFusion effectively mitigates potential degradation in understanding capabilities and surpasses several strong competitors, including Ovis-U1 \citep{wang2025ovis}, Show-o2 \citep{xie2025show}, and Janus-Pro \citep{chen2025janus}.

\begin{figure*}
    \centering
    \vspace{-2em}
    \includegraphics[width=0.95\linewidth]{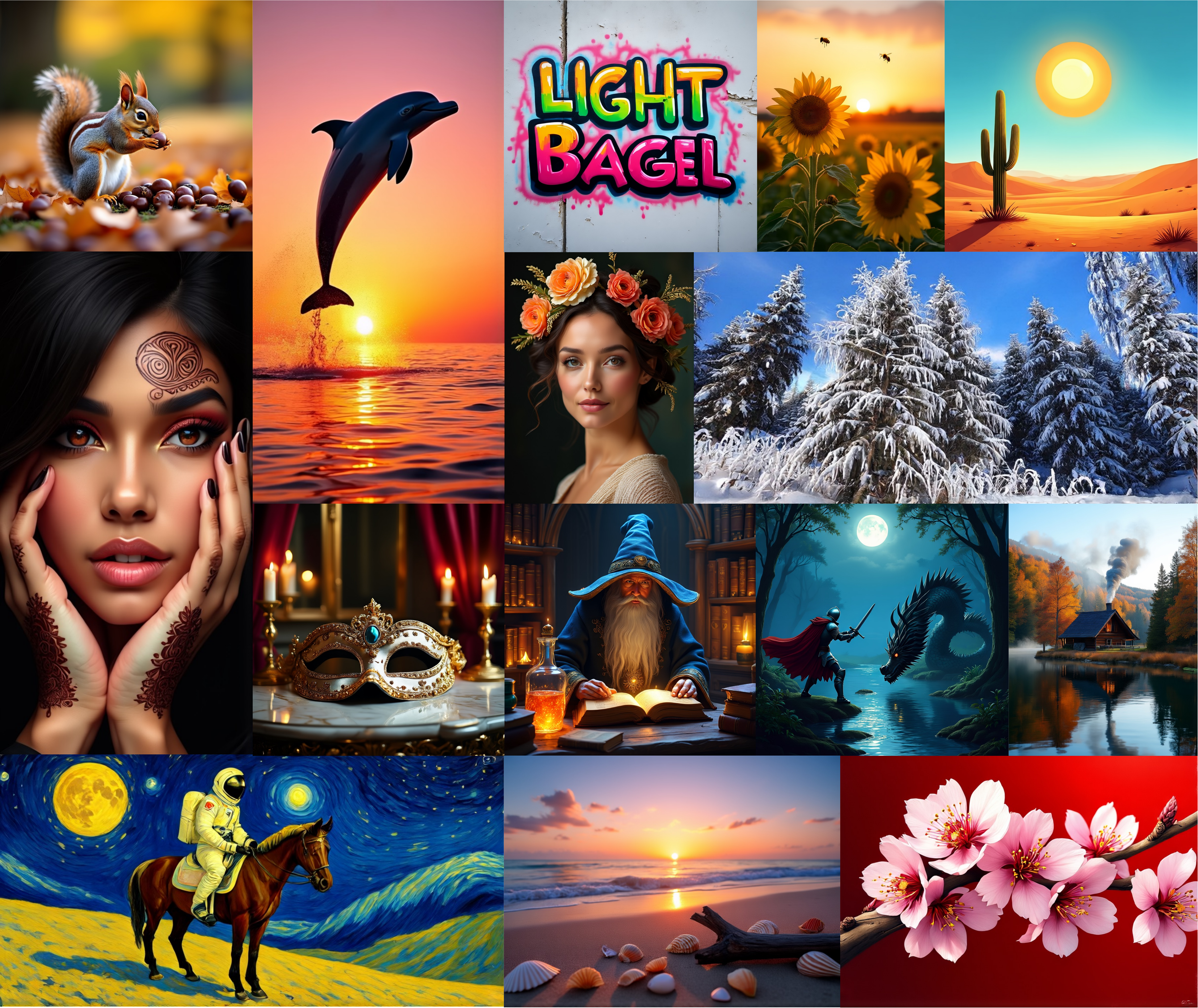}
    \caption{Qualitative text-to-image results from LightFusion, showcasing high-quality generations with strong fidelity to text prompts and consistent rendering across diverse aspect ratios.}
    \label{fig:t2i_vis}
    \vspace{-1em}
\end{figure*}

\subsection{Text-to-Image Generation}
\label{sec:t2i_gen}
We test on widely recognized benchmarks, GenEval and DPG-Bench, to evaluate LightFusion’s text-to-image generation performance. These benchmarks primarily assess the model’s capabilities in compositional image generation and dense prompt following, respectively. In addition, we provide qualitative visualization examples in Figure~\ref{fig:t2i_vis}. Both the quantitative results and qualitative illustrations demonstrate that LightFusion is capable of producing high-fidelity, aesthetically compelling images.

\textbf{GenEval.} As reported in Table \ref{tab:geneval}, LightFusion attains an overall score of 0.91 on GenEval when evaluated with LLM-rewritten prompts, highlighting its strong compositional understanding across diverse generative tasks. This performance surpasses several competitive baselines, including UniPiC (0.86), OmniGen2 (0.86), BAGEL (0.88) and UniPiC2.0 (0.90). Notably, LightFusion is trained with over a order of magnitude of less seen tokens, underscoring its remarkable efficiency in terms of both data and compute resources.

\textbf{DPG-Bench.}
As reported in Table \ref{tab:dpg_bench}, on DPG-Bench, LightFusion achieves an overall score of 82.16, demonstrating competitive performance in long-prompt adherence and complex scene generation. This result surpasses other strong unified models such as BLIP3-o 8B (81.60) and UniWorld-V1 (81.38). Additionally, detailed breakdowns in Table \ref{tab:dpg_bench} indicate that LightFusion maintains consistently strong performance across multiple dimensions of evaluation, including global coherence, entity recognition, attribute understanding, and relational reasoning.

\begin{table}[t]
\centering
\caption{Evaluation of text-to-image generation ability on GenEval benchmark. \dag\ refers to the methods using LLM rewriter.}
\vspace{-.5em}
\resizebox{0.95\linewidth}{!}{
\begin{tabular}{cccccccc}
\toprule
\textbf{Method} & \textbf{Single object}$\uparrow$ & \textbf{Two object}$\uparrow$ & \textbf{Counting}$\uparrow$ & \textbf{Colors}$\uparrow$ & \textbf{Position}$\uparrow$ & \textbf{Color attribution}$\uparrow$ & \textbf{Overall}$\uparrow$\\
\midrule
SDv2.1         & 0.98 & 0.51 & 0.44 & 0.85 & 0.07 & 0.17 & 0.50 \\
SDXL          & 0.96 & 0.74 & 0.39 & 0.85 & 0.15 & 0.23 & 0.55 \\
IF-XL              & 0.97 & 0.74 & 0.66 & 0.81 & 0.13 & 0.35 & 0.61 \\
LUMINA-Next   & 0.92 & 0.46 & 0.48 & 0.70 & 0.09 & 0.13 & 0.46 \\
SD3-medium     & 0.99 & 0.94 & 0.72 & 0.89 & 0.33 & 0.60 & 0.74 \\
FLUX.1-dev    & 0.99 & 0.81 & 0.79 & 0.74 & 0.20 & 0.47 & 0.67 \\
NOVA          & 0.99 & 0.91 & 0.62 & 0.85 & 0.33 & 0.56 & 0.71 \\
OmniGen       & 0.98 & 0.84 & 0.66 & 0.74 & 0.40 & 0.43 & 0.68 \\
\midrule
TokenFlow-XL  & 0.95 & 0.60 & 0.41 & 0.81 & 0.16 & 0.24 & 0.55 \\
Janus         & 0.97 & 0.68 & 0.30 & 0.84 & 0.46 & 0.42 & 0.61 \\
Janus Pro    & 0.99 & 0.89 & 0.59 & 0.90 & 0.79 & 0.66 & 0.80 \\
Emu3-Gen\dag\ & 0.99 & 0.81 & 0.42 & 0.80 & 0.49 & 0.45 & 0.66 \\
Show-o2\dag\        & 1.00 & 0.87 & 0.58 & 0.92 & 0.52 & 0.62 & 0.76 \\
MetaQuery-XL\dag\ & -- & -- & -- & -- & -- & -- & 0.80 \\
UniPic & 0.98 & 0.92 & 0.74 & 0.91 & 0.89 & 0.72 & 0.86 \\
UniPic 2.0\dag\ & -- & -- & -- & -- & -- & -- & 0.90 \\
Ovis-U1\dag\ & 0.98 & 0.98 & 0.90 & 0.92 & 0.79 & 0.75 & 0.89 \\
BAGEL\dag\    & 0.98 & 0.95 & 0.84 & 0.95 & 0.78 & 0.77 & 0.88 \\
OmniGen2\dag       & 0.99 & 0.96 & 0.74 & 0.98 & 0.71 & 0.75 & 0.86 \\
\midrule
BLIP3-o\dag\ 8B  & -- & -- & -- & -- & -- & -- & 0.84 \\
UniWorld-V1\dag\ & 0.98 & 0.93 & 0.81 & 0.89 & 0.74 & 0.71 & 0.84 \\
\rowcolor{blue!10}
LightFusion\dag\  & 1.00 & 0.97 & 0.93 & 0.94 & 0.79 & 0.81 & 0.91 \\
\bottomrule
\end{tabular}
}
\label{tab:geneval}
\end{table}

\begin{table*}[t]
\centering
\caption{Evaluation of text-to-image generation ability on DPG-Bench benchmark.}
\vspace{-.5em}
\resizebox{0.8\linewidth}{!}{
\begin{tabular}{ccccccc}
\toprule
\textbf{Method} & \textbf{Global$\uparrow$} & \textbf{Entity$\uparrow$} & \textbf{Attribute$\uparrow$} & \textbf{Relation$\uparrow$} & \textbf{Other$\uparrow$} & \textbf{Overall$\uparrow$}\\
\midrule
LUMINA-Next      & 82.82 & 88.65 & 86.44 & 80.53 & 81.82 & 74.63 \\
SDXL             & 83.27 & 82.43 & 80.91 & 86.76 & 80.41 & 74.65 \\
PlayGroundv2.5   & 83.06 & 82.59 & 81.20 & 84.08 & 83.50 & 75.47 \\
Hunyuan-DiT      & 84.59 & 80.59 & 88.01 & 74.36 & 86.41 & 78.87 \\
PixArt-$\Sigma$   & 86.89 & 82.89 & 88.94 & 86.59 & 87.68 & 80.54 \\
DALLE3           & 90.97 & 89.61 & 88.39 & 90.58 & 89.83 & 83.50 \\
SD3-medium        & 87.90 & 91.01 & 88.83 & 80.70 & 88.68 & 84.08 \\
FLUX.1-dev       & 82.10 & 89.50 & 88.70 & 91.10 & 89.40 & 84.00 \\
OmniGen          & 87.90 & 88.97 & 88.47 & 87.95 & 83.56 & 81.16 \\
\midrule
TokenFlow-XL     & 78.72 & 79.22 & 81.29 & 85.22 & 71.20 & 73.38 \\
Janus            & 82.33 & 87.38 & 87.70 & 85.46 & 86.41 & 79.68 \\
Janus Pro        & 86.90 & 88.90 & 89.40 & 89.32 & 89.02 & 84.19 \\
Show-o2          & 89.00 & 91.78 & 89.96 & 91.81 & 91.64 & 86.14 \\
EMU3             & 85.21 & 86.68 & 86.84 & 90.22 & 83.15 & 80.60 \\
UniPic           & 89.65 & 87.78 & 90.84 & 91.89 & 91.95 & 85.50 \\
UniPic 2.0       & - & - & - & - & - & 83.79 \\
Ovis-U1             & 82.37 & 90.08 & 88.68 & 93.35 & 85.20 & 83.72 \\
BAGEL            & 88.94 & 90.37 & 91.29 & 90.82 & 88.67 & 85.07 \\
OmniGen2              & 88.81 & 88.83 & 90.18 & 89.37 & 90.27 & 83.57 \\
\midrule
BLIP3-o 8B        & \multicolumn{1}{c}{--} & \multicolumn{1}{c}{--} & \multicolumn{1}{c}{--} & \multicolumn{1}{c}{--} & \multicolumn{1}{c}{--} & 81.60 \\
UniWorld-V1      & 83.64 & 88.39 & 88.44 & 89.27 & 87.22 & 81.38 \\
\rowcolor{blue!10}
LightFusion       & 87.13 & 89.59 & 87.58 & 90.22 & 90.44 & 82.16 \\
\bottomrule
\end{tabular}
}
\label{tab:dpg_bench}
\end{table*}

\begin{figure*}
    \centering
    \includegraphics[width=\linewidth]{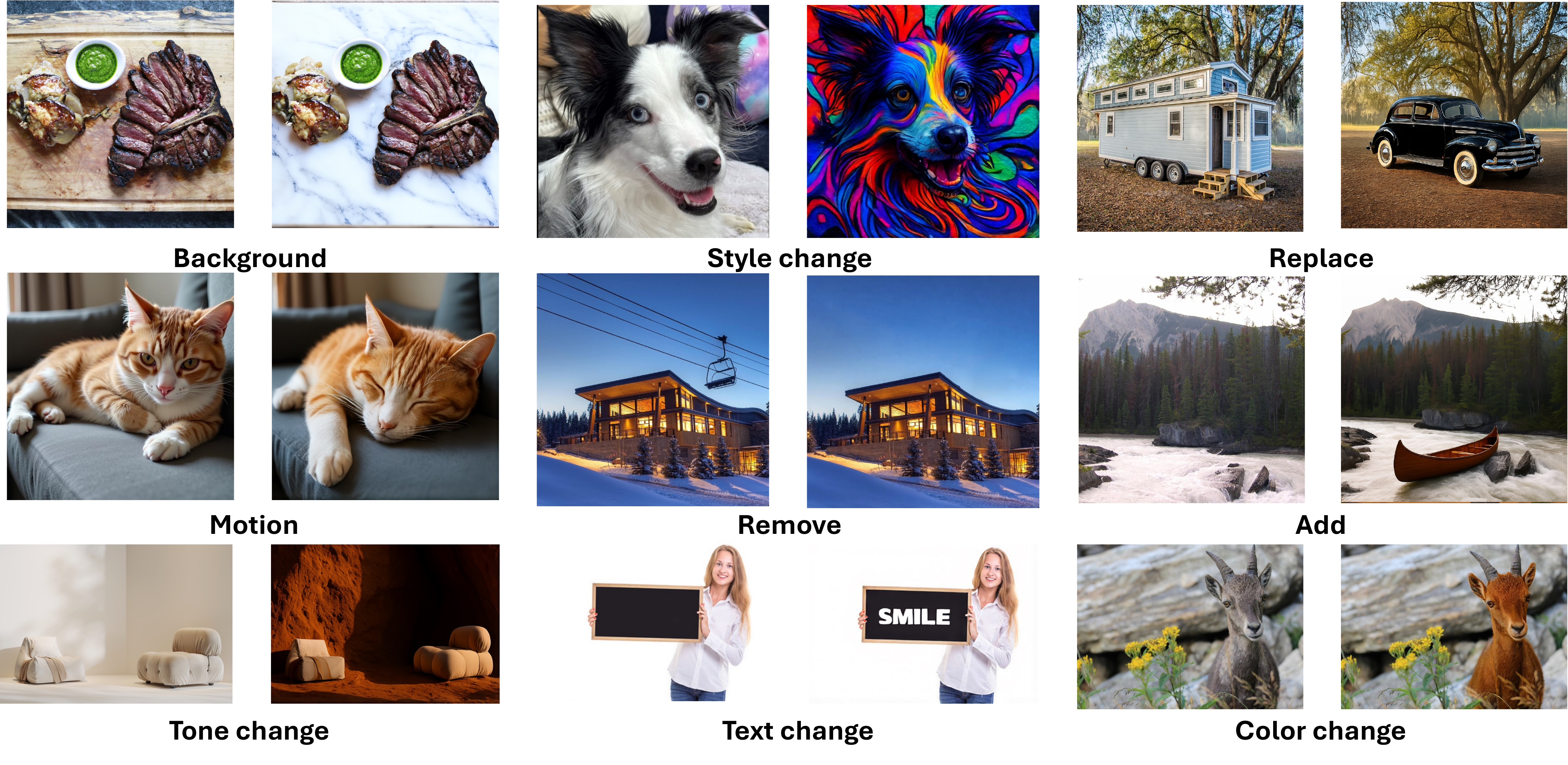}
    \caption{Qualitative image editing results generated by LightFusion. The model exhibits strong instruction following and content preservation capability across a diverse range of editing tasks.}
    \label{fig:edit_vis}
\end{figure*}

\subsection{Image Editing}
\label{sec:image_editing}
We evaluate LightFusion’s image editing capabilities using two widely adopted benchmarks: GEdit-Bench-EN \citep{liu2025step1x} and ImgEdit-Bench \citep{ye2025imgedit}. GEdit-Bench-EN consists of real-world user editing instances, while ImgEdit-Bench encompasses nine distinct editing tasks (e.g., add, remove, alter). We further provide qualitative examples of LightFusion’s editing results in Figure~\ref{fig:edit_vis}. Both quantitative and qualitative results demonstrate that LightFusion delivers strong performance in instruction-based image editing, excelling in both editing accuracy and content preservation.

\textbf{GEdit-Bench-EN.} As shown in Table \ref{tab:gedit}, LightFusion achieves an overall score of 6.06, positioning it among the top-tier unified models. The model demonstrates particular strength in semantic consistency (SC), attaining a score of 6.34, which reflects robust instruction-following capabilities. This result notably surpasses that of UniWorld-V1 (4.93) by a large margin, underscoring the effectiveness of our hybrid ViT+VAE feature fusion strategy compared to only using ViT tokens as condition in image editing task.

\textbf{ImgEdit-Bench.}
As can be seen in Table \ref{tab:imgedit}, LightFusion achieves an overall score of 3.77, outperforming other strong open-source competitors such as UniWorld-V1 (3.26) and OmniGen2 (3.44). Importantly, LightFusion ranks first among open-source models in several key categories, including Add (4.21), Replace (4.55), Remove (3.80), and Hybrid (3.92), which highlights the model’s robust and consistent editing ability across a broad spectrum of tasks.

\begin{table}[t]
\centering
\footnotesize
\caption{Evaluation of image editing ability on GEdit-Bench-EN}
\begin{tabular}{lccc}
\toprule
\textbf{Model} & \textbf{SC} $\uparrow$ & \textbf{PQ} $\uparrow$ & \textbf{Overall} $\uparrow$ \\
\midrule
Gemini-2.0-flash & 6.73 & 6.61 & 6.32 \\
GPT-4o           & 7.85 & 7.62 & 7.53 \\
\midrule
Instruct-Pix2Pix & 3.58 & 5.49 & 3.68 \\
MagicBrush      & 4.68 & 5.66 & 4.52 \\
AnyEdit         & 3.18 & 5.82 & 3.21 \\
ICEdit          & 5.11 & 6.85 & 4.84 \\
Step1X-Edit     & 7.09 & 6.76 & 6.70 \\
\midrule
OmniGen2        & 7.16 & 6.77 & 6.41 \\
BAGEL           & 7.36 & 6.83 & 6.52 \\
Ovis-U1         & --   & --   & 6.42 \\
UniPic          & 6.72 & 6.18 & 5.83 \\
UniPic 2.0      & -- & -- & 7.10 \\
\midrule
UniWorld-V1     & 4.93 & 7.43 & 4.85 \\
\rowcolor{blue!10}
LightFusion      & 6.34 & 7.31 & 6.06 \\
\bottomrule
\end{tabular}
\label{tab:gedit}
\end{table}

\begin{table*}[t]
\centering
\caption{Evaluation of image editing ability on ImgEdit-Bench.}
\resizebox{\linewidth}{!}{
\begin{tabular}{lccccccccc c}
\toprule
\textbf{Model} & \textbf{Add} & \textbf{Adjust} & \textbf{Extract} & \textbf{Replace} & \textbf{Remove} & \textbf{Background} & \textbf{Style} & \textbf{Hybrid} & \textbf{Action} & \textbf{Overall} \\
\midrule
GPT-4o & 4.61 & 4.33 & 2.90 & 4.35 & 3.66 & 4.57 & 4.93 & 3.96 & 4.89 & 4.20 \\
\midrule
MagicBrush     & 2.84 & 1.58 & 1.51 & 1.97 & 1.58 & 1.75 & 2.38 & 1.62 & 1.22 & 1.90 \\
Instruct-Pix2Pix & 2.45 & 1.83 & 1.41 & 2.01 & 1.44 & 1.44 & 3.55 & 1.20 & 1.46 & 1.88 \\
AnyEdit        & 3.18 & 2.95 & 1.14 & 2.49 & 2.21 & 2.88 & 3.82 & 1.56 & 2.65 & 2.45 \\
UltraEdit      & 3.44 & 2.81 & 2.00 & 2.96 & 2.45 & 2.83 & 3.76 & 1.91 & 2.98 & 2.70 \\
Step1X-Edit    & 3.88 & 3.41 & 1.76 & 3.40 & 2.83 & 3.16 & 6.63 & 2.52 & 2.52 & 3.06 \\
ICEdit         & 3.58 & 3.39 & 1.73 & 3.15 & 2.93 & 3.08 & 3.84 & 2.04 & 3.68 & 3.05 \\
\midrule
OmniGen2     & 3.74 & 3.54 & 1.77 & 3.21 & 2.77 & 3.57 & 4.81 & 2.30 & 4.14 & 3.43 \\
BAGEL        & 3.56 & 3.31 & 1.88 & 2.62 & 2.88 & 3.44 & 4.49 & 2.38 & 4.17 & 3.20 \\
Ovis-U1      & 4.12 & 3.92 & 2.36 & 4.09 & 3.57 & 4.22 & 4.69 & 3.23 & 3.61 & 3.98 \\
UniPic       & 3.66 & 3.51 & 2.06 & 4.31 & 2.77 & 3.77 & 4.76 & 2.56 & 4.04 & 3.49 \\
UniPic 2.0   & - & - & - & - & - & - & - & - & - & 4.06 \\
\midrule
UniWorld-V1    & 3.82 & 3.66 & 2.31 & 3.45 & 3.02 & 2.99 & 4.71 & 2.96 & 2.74 & 3.26 \\
\rowcolor{blue!10}
LightFusion   & 4.21 & 3.23 & 1.83 & 4.55 & 3.80 & 4.15 & 4.66 & 3.93 & 3.60 & 3.77 \\
\bottomrule
\end{tabular}
}
\label{tab:imgedit}
\end{table*}

\begin{figure*}[!t]
    \centering
    \begin{minipage}[b]{1.0\textwidth}
        \centering
        \begin{subfigure}[b]{0.48\textwidth}
        
            \centering

            \includegraphics[height=4.5cm]{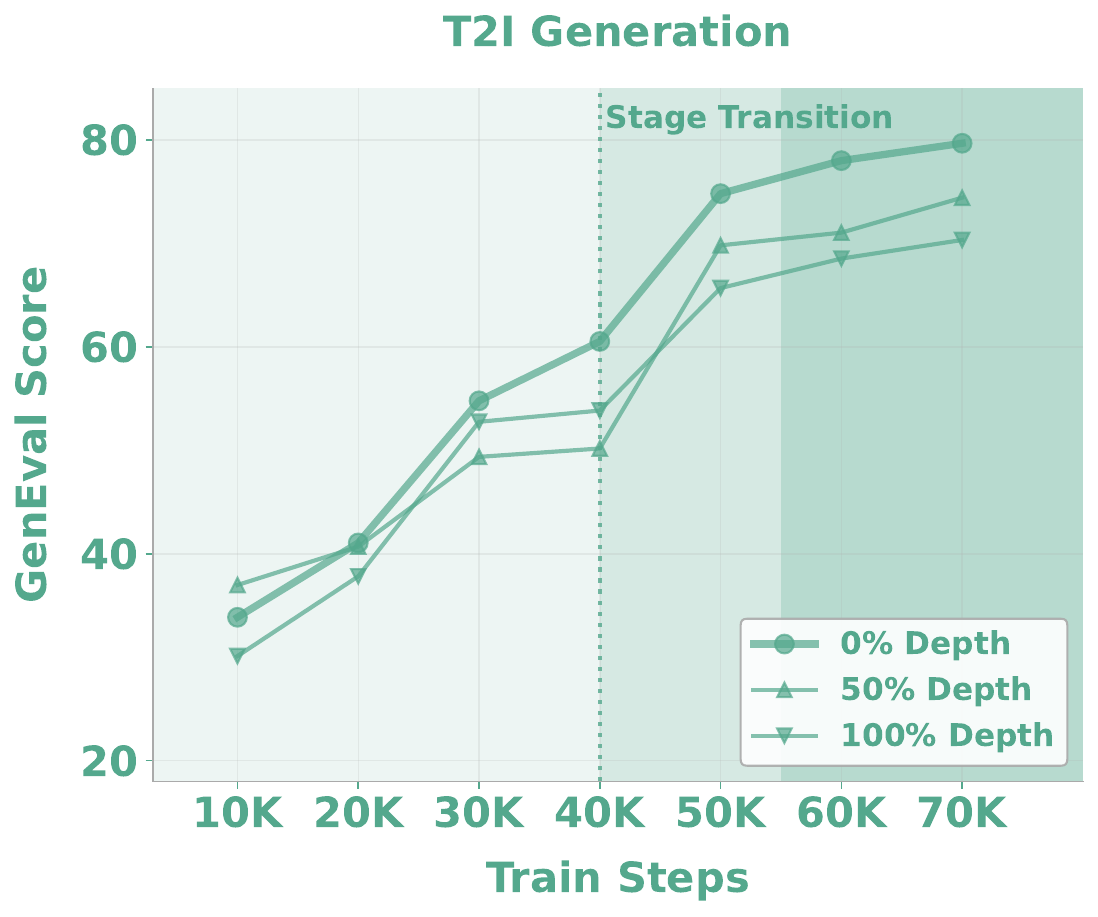}
            \label{fig:scaling_geneval}
        \end{subfigure}
        \hfill
        \begin{subfigure}[b]{0.48\textwidth}
            \centering

            \includegraphics[height=4.5cm]{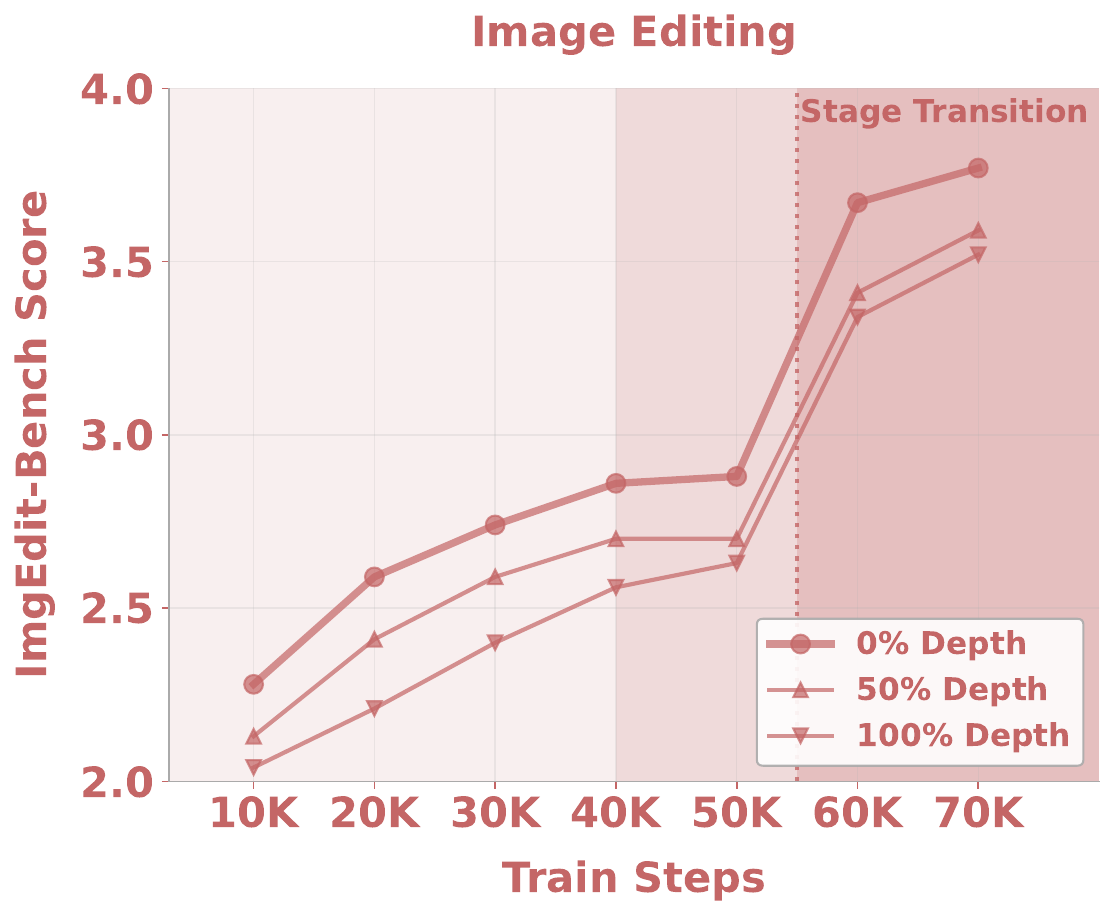}
            \label{fig:scaling_imgedit}
        \end{subfigure}
    \end{minipage}
    \caption{Deep fusion \vs shallow fusion design choices. Regions of different colors represent different training stages. The “0\% Depth” deep fusion approach in our LightFusion consistently outperforms other options.}
    \vspace{-1em}
    \label{fig:deep_vs_shallow_fusion}
\end{figure*}

\begin{table*}[t]
\centering
\caption{Ablation study on visual tokenizer and timestep shift choices.}
\begin{subtable}[h]{0.45\textwidth}
\centering
\small
    \begin{tabular}{c|cc}
    \shline
    Encoder   &GEdit-EN &ImgEdit \\
    \shline
    ViT  & 3.91 & 2.65 \\
    VAE  & 4.93 & 3.38 \\
    \rowcolor{blue!10}
    ViT + VAE & 5.61 & 3.57 \\ 
    \shline
    \end{tabular}
    \caption{Evaluation of different visual tokenizer choices.}
    \label{tab:visual_tok}
\end{subtable}
\hspace{2em}
\begin{subtable}[h]{0.45\textwidth}
\centering
\scriptsize
    \begin{tabular}{c|cccc}
    \shline
    Timestep Shift   &DPG-Bench &ImgEdit \\
    \shline
    1  & 76.67 & 3.07 \\
    2  & 78.84 & 3.36 \\
    \rowcolor{blue!10}
    4  & 81.77 & 3.57 \\ 
    \shline
    \end{tabular}
    \caption{Evaluation of different timestep shift choices.}
    \label{tab:timestep_shift}
\end{subtable}
\label{table:design choices}
\vspace{-1em}
\end{table*}

\subsection{Ablation Study}
\label{sec:ablation_study}
For ablation study experiments, we keep the general training setup for the \textit{"Deep Fusion vs Shallow Fusion"} study, while opt for 40K training steps for the others for faster experiment cycles.

\textbf{Deep Fusion vs Shallow Fusion.} Previous efficient UMM approaches typically employ a lightweight connector that maps the final output of the understanding branch as a conditional input to the generation branch. In contrast, our Double Fusion design allows language and visual tokens to interact from the earliest layers, enabling deeper and more continuous cross-modal integration.

To systematically compare the two strategies, we vary the depth at which features from the VLM are injected into the generation pathway. Specifically, “0\% Depth” denotes the case where the $i$-th DiT block is conditioned on the $i$-th VLM block’s output. When the VLM blocks are exhausted, the final VLM output is repeated as input for the remaining DiT blocks. Conversely, “100\% Depth” corresponds to conditioning all DiT blocks exclusively on the final VLM output, effectively repeating it across all multimodal self-attention layers. Importantly, we keep the total number of multimodal self-attention layers fixed, ensuring identical parameter counts and a fair comparison.

As shown in Figure~\ref{fig:deep_vs_shallow_fusion}, the “0\% Depth” option—adopted in our LightFusion model—consistently outperforms shallow or early-layer fusion for both text-to-image and image-editing tasks. We attribute this advantage to the fact that the final VLM representations encode high-level semantics more suitable for next-token prediction rather than multimodal alignment.

\textbf{Visual Tokenizer Choices.} VAE and ViT encoders provide complementary visual representations: VAEs emphasize low-level details, while ViTs capture high-level semantic information. To assess which type of information is more suitable for image editing, we conduct an ablation study on visual tokenizer choice, with results reported in Table~\ref{tab:visual_tok}. The findings indicate that both sources of information are essential—combining low-level details with high-level semantics leads to more consistent and accurate image editing outcomes.

\textbf{Training Timestep Shift.} Shifting timesteps during inference has been shown to improve generation quality in state-of-the-art models~\citep{flux1dev,wan2025}. In our preliminary experiments, we observe that the strong base DiT pathway already demonstrates robust denoising capabilities. Building on this, and following the inference-time timestep shifting strategy~\citep{flux1dev,wan2025}, we increase the training diffusion timestep range from 1.0 to 4.0 to achieve more noisy corrupted samples. As reported in Table~\ref{tab:timestep_shift}, a timestep shift larger than 1 consistently lead to better results.

\section{Conclusion}
This work introduces LightFusion, a unified multimodal model for both understanding and generation that achieves state-of-the-art performance across diverse tasks while requiring substantially fewer training tokens and compute than prior leading UMMs. These advantages are largely benefited by our Double Fusion design, which enables effective cross-modal feature interactions and naturally integrates high-level semantics with low-level visual details. By releasing the entire suite of code, models, and data, we hope to support reproducibility and accelerate progress in unified multimodal modeling.

\bibliography{iclr2026_conference}
\bibliographystyle{iclr2026_conference}


\end{document}